%% file: main.tex
\documentclass[sigconf]{acmart}

\usepackage{amsmath}

\usepackage[linesnumbered, ruled, vlined]{algorithm2e}

\author{Hanqi Chen}
\affiliation{%
  \institution{Shanghai Jiao Tong University}
  \department{SJTU Paris Elite Institute of Technology}
  \city{Shanghai}
  \country{China}}
\email{caffinities@sjtu.edu.cn}

\author{Zhongyin Zhao}
\affiliation{%
  \institution{Shanghai Jiao Tong University}
  \city{Shanghai}
  \country{China}}
\email{zhao_zhongyin@sjtu.edu.cn}

\author{Ye Chen}
\affiliation{%
  \institution{Shanghai Jiao Tong University}
  \city{Shanghai}
  \country{China}}
\email{chenye123@sjtu.edu.cn}

\author{Zhujin Liang}
\affiliation{%
  \institution{PhiGent Robotics}
  \city{Beijing}
  \country{China}}
\email{zhujin.liang@phigent.ai}

\author{Bingbing Ni}
\authornote{Corresponding author.}
\affiliation{%
  \institution{Shanghai Jiao Tong University}
  \city{Shanghai}
  \country{China}}
\email{nibingbing@sjtu.edu.cn}

\acmSubmissionID{mfp3562}

\copyrightyear{2025}
\acmYear{2025}
\setcopyright{acmlicensed}
\acmConference[MM '25] {Proceedings of the 33rd ACM International Conference on Multimedia}{October 27--31, 2025}{Dublin, Ireland.}
\acmBooktitle{Proceedings of the 33rd ACM International Conference on Multimedia (MM '25), October 27--31, 2025, Dublin, Ireland}
\acmISBN{979-8-4007-2035-2/2025/10}
\acmDOI{10.1145/3746027.3755392}

\begin{document}

\title{SVGThinker: Instruction-Aligned and Reasoning-Driven Text-to-SVG Generation}

\begin{abstract}
Scalable Vector Graphics (SVG) is a code structure used to represent visual information, and with the powerful capabilities of large language models, it holds significant research potential. Current text-to-SVG generation methods lack generalization capabilities and struggle with accurately adhering to input generation instructions. In this paper, we propose a novel approach for generating SVG using large language models, named SVGThinker, which incorporates a reasoning process to align the generation of SVG code with the visualization process, while supporting all SVG primitives. Through sequential rendering of SVG primitives, we first use a multimodal model to annotate the SVG, followed by sequential updates corresponding to the incremental additions of primitives. We then employ a supervised training framework based on Chain-of-Thought reasoning, which enhances the model’s robustness and reduces the risk of errors or hallucinations. Through comparisons with state-of-the-art baseline models, our experiments show that our model generates more stable, high-quality, and editable SVG code. In contrast to image-based methods, our approach preserves the structural advantages of SVG and supports precise, hierarchical editing. We believe our work opens new directions for SVG generation, with potential applications in design, content creation, and automated SVG-based graphic generation.
\end{abstract}



\begin{teaserfigure}
  \includegraphics[width=\textwidth]{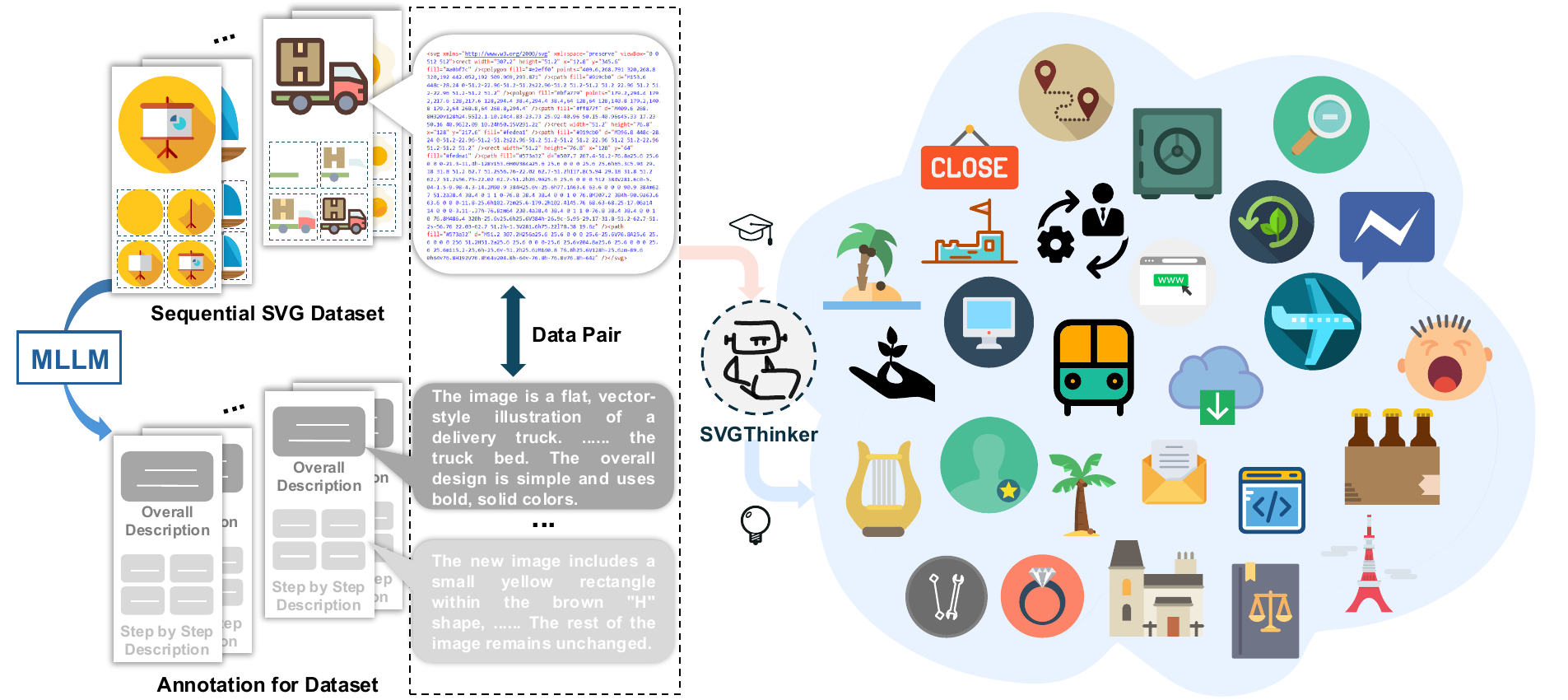}
  \caption{SVGThinker, our proposed text-to-SVG reasoning model, leverages Multimodal Large Language Model (MLLM) to construct sequential SVG textual annotations, creating both overall descriptions and step-by-step descriptions of visual changes in SVG primitives. By using this paired data, we enable the reasoning model to understand SVG primitives and their corresponding visual semantic information, thereby generating SVGs aligned with the instructions. Through modifying generation instructions, SVGThinker allows for precise and controllable editing of SVGs, showcasing capabilities that were previously unattainable by other models.}
  \Description{Task of SVGThinker.}
  \label{fig:teaser}
\end{teaserfigure}


\maketitle

\section{Introduction}

Scalable Vector Graphics (SVG) is a powerful and widely used vector graphics format. It provides a flexible and scalable method for creating, editing, and distributing images through XML-based textual descriptions. Compared to raster images, SVG maintains clarity and sharpness at any resolution while occupying minimal storage space. These advantages make SVG highly applicable across various fields, particularly in design, art, creativity, and industrial applications. However, despite its widespread use, understanding its syntax or mastering its creation methods poses significant challenges for humans. The source code of SVG and the visual image representation and semantic information it encapsulates are complex, making it difficult for humans to comprehend. In recent years, generative models based on deep learning techniques~\cite{carlier2020deepsvg} have partially addressed the difficulty of creating and using SVGs by facilitating the learning and understanding process. These methods can be categorized into two paradigms.

The first paradigm is optimization-based methods, which typically include image generation models~\cite{rombach2022high} and differentiable SVG renderers~\cite{Li:2020:DVG}, such as VectorFusion~\cite{jain2023vectorfusion}, SVGDreamer~\cite{xing2024svgdreamer}, etc. This approach uses images as visual targets and optimizes the similarity between the SVG and the original image. These methods typically represent images using paths, and often produce visual artifacts, which diminish some of the expressive advantages of SVGs, and generate redundant and chaotic primitives that cannot be easily edited or modified later. This approach leads to a loss of semantic understanding of the SVG code, creating a bottleneck for further optimization in generating high-quality SVGs.

The second paradigm is based on generative models~\cite{wu2023iconshop}, typically using image generation models~\cite{flux2024} or autoregressive models to generate SVGs in a feedforward manner. Methods such as LayerTracer~\cite{song2025layertracer} use generated SVG-style images, which are then vectorized using tools like VTracer~\cite{vtracer}. This approach also suffers from content redundancy, image flaws, and issues with unreadable or uneditable SVG source code. Another approach is to use autoregressive models based directly in the SVG source code space, like IconShop~\cite{wu2023iconshop}, or to use language models like StarVector~\cite{rodriguez2023starvector}, or GPT-4o~\cite{achiam2023gpt}. These methods either limit the range of SVG primitives using their own SVG tokenizer, which weakens their expressive capabilities and fails to properly understand the language input information~\cite{wu2023iconshop}, or struggle to capture the semantic information contained in it due to the abstract nature of SVG source code representation~\cite{achiam2023gpt}, leading to challenges in instruction-aligned generalization and editing. 

To address these challenges, we propose SVGThinker, a large language model-based text-to-SVG generation model, which supports all SVG primitives. Our approach is motivated by the inherent tree structure and visual-semantic information of SVG codes. Specifically, in contrast to general LLM-based pipelines which directly pair textual descriptions and SVG codes for training~\cite{xing2024empowering}, we redesigned a novel textual annotation process based on parsing and reconstructing the tree structure of SVGs, ensuring a clear correspondence between the visual meaning and textual description corresponding to each SVG primitive. At the same time, we train a LLM integrated with a Chain-of-Thought (CoT)~\cite{wei2022chain} reasoning process aligned with human cognitive understanding and the logical construction of SVG instructions. Through the proposed method, we make the generated SVG representations more concise and meet the high standards of human-made SVGs, while adhering to the way humans conceptualize SVGs.

Extensive comparisons with existing state-of-the-art models demonstrate that our method surpasses previous approaches by generating SVGs that are more aligned with textual descriptions, more readable, and more reasonable. Additionally, based on the input textual instructions, fine-grained editing is possible. Our experiments show that by accurately modifying parts of the input prompt, such as position, color, elements, and structure, the model can precisely adjust the generated SVG code and the corresponding visual output. This capability, derived from the model’s reasoning ability and the data annotation from prior work, is something that no previous method has been able to achieve.

\section{Related Works}

\subsection{SVG Generation}

Early SVG generation methods were primarily based on image vectorization techniques. After significant advancements in deep learning, SVG-VAE~\cite{lopes2019learned}, based on variational autoencoders, began conditional generation of SVGs. Subsequently, methods like VectorFusion~\cite{jain2023vectorfusion} and SVGDreamer~\cite{xing2024svgdreamer} emerged, combining image generation models and vectorization techniques to achieve higher-quality SVG generation. Additionally, many related SVG generation works~\cite{zhao2024vector} based on diffusion models~\cite{ho2020denoising} have been proposed. As autoregressive generation models~\cite{brown2020language, achiam2023gpt} demonstrated strong generalization capabilities, IconShop~\cite{wu2023iconshop} introduced an SVG tokenization approach that constrains the range and expression based on the structure of SVG instructions. Recent trends have further leveraged large language model-based methods, such as StarVector~\cite{rodriguez2023starvector} and Chat2SVG~\cite{wu2024chat2svg}. These approaches exhibit higher potential and theoretically complete SVG support. However, existing methods have not explored the corresponding semantic information of SVG code, which limits their performance, motivating the development of SVGThinker.

\subsection{Large Language Models}

Large language models, such as GPT-4~\cite{achiam2023gpt}, have revolutionized natural language processing and gradually expanded to other modalities. For example, through methods like LLaVA~\cite{liu2023visual}, image modality support has been integrated, further advancing large language models toward general intelligence. By increasing the model parameters and using transformer~\cite{vaswani2017attention} architectures, language models have demonstrated scaling laws~\cite{kaplan2020scaling}, and their high performance is becoming a foundational model~\cite{radford2019language} for other tasks, such as SVG generation. As model parameters have reached certain bottlenecks, the focus of language models has gradually shifted to reasoning applications and more powerful task-solving capabilities~\cite{zhang2023multimodal}. Through reinforcement learning, supervised learning, and new paradigms such as CoT~\cite{wei2022chain}, model performance has been further enhanced and aligns more closely with human thinking processes for corresponding tasks. The work presented in this paper also relies on the improved understanding and learning capabilities of models, enabling the model’s reasoning process to align with the SVG construction process, thereby demonstrating precise editing capabilities for the first time.

\input{new_methods}

\input{experiments}

\section{Conclusion}

We introduce SVGThinker, a reasoning-driven framework for SVG generation that bridges the gap between textual instructions and visual output through a sequential and hierarchical generation process. Our model leverages large language models and a multimodal approach to ensure precise alignment between SVG code and its corresponding visual representation, while supporting all SVG primitives. By employing Chain-of-Thought reasoning and a structured annotation process, we enhance the model’s robustness and accuracy in generating stable, high-quality, and editable SVG code. Our experiments show that SVGThinker outperforms state-of-the-art methods in generation quality, editing flexibility, and alignment with textual instructions. Additionally, our user study indicates a preference for our method in terms of ease of use, aesthetic quality, and instruction adherence.

\begin{acks}
This work is supported by the Science and Technology Commission of Shanghai Municipality under research grant No. 25ZR1401187.
\end{acks}

\bibliographystyle{ACM-Reference-Format}
\balance
\bibliography{references_local} 

\appendix

\end{document}

%% file: new_methods.tex
\section{Methodology}

\begin{figure*}
    \centering
    \includegraphics[width=0.96\linewidth]{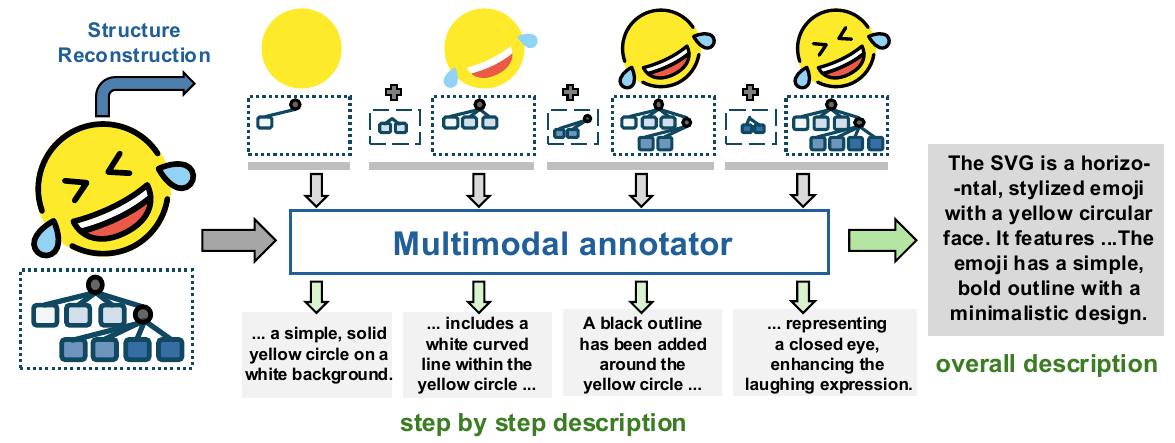}
    \caption{The process of sequential textual annotation for SVG. First, we perform an overall description of the SVG. Then, by parsing and reconstructing the SVG structure, we obtain a set of textual descriptions related to the sequence of SVG primitives, based on the visual changes at each step.}
    \label{fig:annote_pipe}
\end{figure*}

In this section, we describe the entire process of dataset creation and the corresponding model training workflow. First, we collect the data, followed by preprocessing the SVG data. We then annotate the SVG using a multimodal large language model, and subsequently train the large language model based on reasoning logic.

\subsection{Data Preparation for Instruction Alignment}

\noindent\textbf{Data Collection and Curation. } We collected a dataset on the Kaggle platform~\cite{svgkaggle} consisting of 5,269 fine-grained categories, totaling more than 300,000 icon samples. From the SVGRepo platform~\cite{svgrepo}, we gathered more than 100,000 vector graphic data points through classified retrieval, organizing them into 34 coarse-grained semantic categories. The semantic matching accuracy of these data was manually calibrated to ensure comprehensive semantic coverage. Subsequently, we used the tools provided by SVGO to optimize all the SVGs and removed any samples with excessively long SVG.

\noindent\textbf{SVG structure reconstruction for image generation.} After completing the optimization step for the SVG files, we proceeded to generate intermediate-step images that are aligned with the instructions. To achieve this, we parsed the SVG files according to their hierarchical structure. The drawing process in SVG is executed directly in the order of instructions. However, since SVG instructions are structured in an XML-like tree format, they are not processed in a simple sequential manner. In addition, many structures have attributes that affect the child nodes. To better align the SVG code with the rendered images, we developed a simple algorithm for reconstructing the SVG structure based on the visual changes observed during image rendering.

Through this approach, for each individual SVG file, we generate a series of images that guide the reconstruction process. The updates between each image are aligned with the corresponding instructions, helping the language model learn the relationships between SVG code and its visual output.

\noindent\textbf{SVG annotation.} We employ the LLaVA framework~\cite{liu2023visual} for image-flow annotations and filtering using CLIP~\cite{radford2021learning} and Perplexity. When annotating SVG-rendered images, we face several challenges. On one hand, the correspondence between SVG code and image elements is difficult to understand. On the other hand, the sequential rendering of SVG elements results in certain layering effects, which complicate matters for the language model. To address these challenges, we utilize the image sequences generated from the earlier structured SVG rendering to assist in identifying occlusions, element correspondences, and textual descriptions of the images.

Previously, when annotating SVG images, the annotation was typically performed only on the rendered image once~\cite{wu2023iconshop, xing2024empowering, rodriguez2023starvector}. This method did not optimize the structure of the SVG itself, failing to reflect the advantages of the SVG format in generating visual images. Furthermore, it could not stably reconstruct the SVG image from text alone. Moreover, the previous annotation approach lacked robustness, offering both detailed descriptions with precise location guidance as well as vague summaries and category recognition of the SVG, which caused inconsistencies. This inconsistency hindered the editability and stability of the generation process and negatively impacted model training.

To obtain serialized SVG-rendered image annotations, we used a multimodal large language model to annotate sequences of images generated from an SVG. This provides precise annotation guidance that aligns with human understanding and supports multi-round image-text dialogues for comparing image differences. For the annotation process, we first present a complete SVG-rendered image and have the multimodal model generate a detailed description, denoted as $t_g$. Then, in a multi-round dialogue, we annotate each step of SVG construction by adding one instruction at a time. With each new image added, the model is asked to describe the changes from the previous image, thus guiding the next stage of SVG generation.

Assume we have an image sequence $(I_1, I_2, \dots, I_n), n\in\mathbb{N}^*$, from the rendering of the first SVG instruction to the final completion of the SVG rendering. For each input image $I_i$, we obtain a description of the differences from the multimodal model by Equation~\ref{eq:sample}, where $I_n$ denotes the complete SVG image.

\begin{align}
\label{eq:sample}
    t_i \sim \mathbb{P}(x \mid I_i, I_{i-1}&, t_g, I_n)\quad \forall i \in \{1,\dots,n-1\}
\end{align}

In this way, we derive the text sequence $t_{n-1}$ for each difference description. Throughout this dialogue, since we always retain the initial full annotation of the complete SVG, the textual descriptions of the gradually constructed SVG will not be subject to errors or hallucinations due to the incompleteness of the SVG at later stages. Through this session-based approach, we ensure the accuracy of every textual annotation step, making it more robust compared to the traditional method of annotating with only two rendered images (before and after). By retaining the initial comprehensive annotation of the complete SVG, subsequent annotations of the SVG as it is incrementally constructed maintain accuracy. This approach is more robust compared to annotating just the two images before and after changes in the rendering process. The overall process of text annotation is shown in Table \ref{fig:annote_pipe}.

We employed InternVL2.5-38B-MPO-AWQ open-source model~\cite{chen2024expanding} for this data annotation process. This model is highly efficient due to its int4 quantization~\cite{lin2024awq}, and we utilized the LMDeploy tool~\cite{2023lmdeploy} to efficiently complete the entire data annotation task.

\subsection{Training of LLM for Reasoning-Driven SVG Generation}

\begin{figure*}
    \centering
    \includegraphics[width=\linewidth]{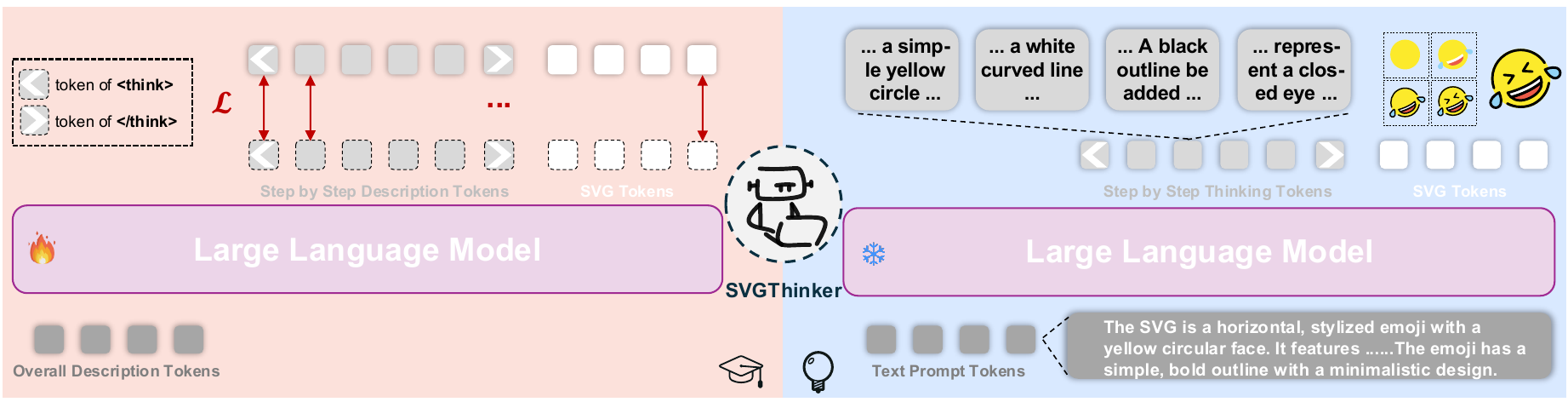}
    \caption{The training and inference pipeline of SVGThinker. On the left is the training process, where we use overall descriptions to train the model to reason through the process of constructing an SVG, incorporating special thinking tokens for distinction. On the right is the inference process, where our model receives user instructions, analyzes the instructions, conceptualizes the corresponding SVG primitives, and then generates the SVG code based on the reasoning process.}
    \label{fig:main_pipe}
\end{figure*}

Our goal is to align the reasoning process of the SVG generation model with the inherent sequential drawing instructions of SVGs, proceeding layer-by-layer according to the canvas-building process. Similar to typical large language models, our model takes natural language descriptions and commands to generate SVG code. It follows user instructions by internally formulating a sequence of drawing steps, explicitly presenting this reasoning process before generating the final SVG code. The training targets are paired textual annotations derived from comparisons of SVG generation sequences and their corresponding SVG outputs. Inputs consist of comprehensive SVG descriptions produced by a multimodal model. This approach tightly integrates the training procedure, data annotation workflow, and downstream user scenarios.

Assume that we are given an SVG composed of a sequence of primitive instructions $(s_1, s_2, \dots, s_{n}), n\in\mathbb{N}^{*}$. Based on our previously described annotation pipeline, we obtain a complete textual description of the SVG, denoted $t_g$, along with an ordered sequence of intermediate descriptions $(t_1, t_2, \dots, t_{n-1})$.

During the training of the reasoning model, we begin by modeling  probability distribution in Equation~\ref{eq:proba}.

\begin{equation}
\label{eq:proba}
    \mathbb{P}( (t_1, t_2, \dots, t_{n-1}) \mid t_g) = \prod^{n-1}_{i=1}\mathbb{P}(t_i\mid (t_{j})_{\forall j<i}, t_g)
\end{equation}

Subsequently, based on the inferred drawing process - which corresponds to the ordered sequence of intermediate descriptions - we derive a refined probability distribution to optimize the model in Equation~\ref{eq:second_proba}. 
\begin{align}
    \mathbb{P}((s_1, s_2, \dots, s_n) \mid (t_1, t_2, \dots, t_{n-1}), t_g) \notag \\
    \label{eq:second_proba}
    = \prod_{i=1}^{n} \mathbb{P}\left(t_i  \mid (s_{j})_{\forall j<i}, (t_1, t_2, \dots, t_{n-1}), t_g\right)
\end{align}

The learning objective is then grounded in the autoregressive loss function used in language generation models. Specifically, we minimize the negative log-likelihood of the target sequence conditioned on the previous steps in Equation~\ref{eq:loss}, where $(x_1, \dots, x_n)$ is the target sequence.

\begin{align}
\label{eq:loss}
    \mathcal{L} = -\sum_{i=1}^{n} \log & P(x_i \mid x_1, \dots, x_{i-1}) 
\end{align}

Through this training framework in Figure~\ref{fig:main_pipe}, the model first learns the structured drawing process of SVGs, acquiring an understanding of how intermediate visual steps unfold. Based on this reasoning process, the model then generates the corresponding SVG instructions, thereby achieving our cognitively aligned objective for structured SVG generation.

%% file: experiments.tex
\section{Experiments}

In this section, we present our training procedure, baseline comparisons, ablation studies, and a user study evaluating subjective experiences. We highlight our model’s precise instruction editing and compare it with state-of-the-art methods. To evaluate our approach, we use the same Qwen2.5-7B architecture~\cite{qwen2.5}. Specifically, we train our model initialized from the chain-of-thought~\cite{wei2022chain} distilled weights provided by DeepSeek~\cite{guo2025deepseek}, and compare it against a baseline trained directly on text-labeling and SVG pairs using original Qwen weights~\cite{qwen2.5}.

\begin{table*}
  \caption{Quantitative comparison with SOTA models. We also compared the file size, the number of primitives used, and the variety of supported primitives. Our method supports all primitive types, generating higher-quality, more efficient, and compact SVGs with a more concise use of primitives.}
  \label{tab:metrics_comparison}
  \begin{tabular}{c|c|c|c|c|c|c}
    \toprule
    Methods & FID$\downarrow$ & CLIP Score$\uparrow$ & FID-CLIP$\downarrow$ & Primitives Support & File Size (KB) & Primitives Used\\
    \midrule
    LayerTracer~\cite{song2025layertracer} & 54.75 & 0.2290 & 30.46 & path & 16.25 & 17.83 \\
    SVGDreamer~\cite{xing2024svgdreamer} & 240.87 & 0.1923 & 150.34 & path & 282.13 & 513.0 \\
    IconShop~\cite{wu2023iconshop} & 89.24 & 0.2672 & 53.79 & path & 3.14 & 1.042 \\
    GPT-4o-2024-11-20~\cite{achiam2023gpt}&62.56& 0.1715&43.93& all&0.67&5.62 \\
    DeepSeek-R1~\cite{guo2025deepseek}&153.04&0.1160&111.42&all&0.71&5.30\\
    SVGThinker & \textbf{34.06} & \textbf{0.2765} & \textbf{21.08} & all & 1.16 & 3.707 \\
    \bottomrule
  \end{tabular}
\end{table*}
\subsection{Experiments Settings}

\noindent\textbf{Dataset.} The training data used in this study is based on the dataset we previously constructed. From the 270,436 samples, we randomly selected 1,000 data pairs as the test set, ensuring no overlap with the training data. Evaluation and generation were performed on these prompts, guaranteeing that the model had no prior knowledge of them.

\noindent\textbf{Training settings.} This setup enables direct comparison between the two models to verify the reliability of our method. Both models were trained using the AdamW optimizer~\cite{loshchilov2017decoupled} with a learning rate of $4\times10^{-5}$ and cosine annealing~\cite{loshchilov2016sgdr}, with a minimum learning rate of $2 \times 10^{-6}$. One epoch was used for warmup, followed by 9 training epochs. We utilized the Accelerate library and LLaMA-Factory~\cite{zheng2024llamafactory} for training acceleration, and LMDeploy~\cite{2023lmdeploy} for efficient inference and sampling. All experiments were run on eight A800 GPUs.

\noindent\textbf{Baselines.} For baseline comparison, we included both commercial closed-source large models and strong open-source SVG-specialized generative models. Among the former, GPT-4o-2024-11-20~\cite{achiam2023gpt} and DeepSeek-R1~\cite{guo2025deepseek} were evaluated via their official APIs. For open-source models, we selected SVGDreamer~\cite{xing2024svgdreamer}, which uses image generation and differentiable SVG rendering, IconShop~\cite{wu2023iconshop}, an autoregressive SVG generation model, and LayerTracer~\cite{song2025layertracer}, which generates icon-style images and vectorizes them into SVG format.

\noindent\textbf{Metrics.} We evaluated SVG generation quality using three metrics: FID~\cite{heusel2017gans}, CLIP~\cite{radford2021learning}, and FID-CLIP. FID measures the distance between the distributions of generated and real SVGs using Inception-v3~\cite{szegedy2016rethinking}, while CLIP measures alignment between text and image via cosine similarity. FID-CLIP uses CLIP to compute the image feature vectors for distribution comparison.

\subsection{Experimental Results and Discussion}

\begin{figure*}
    \centering
    \includegraphics[width=0.92\linewidth]{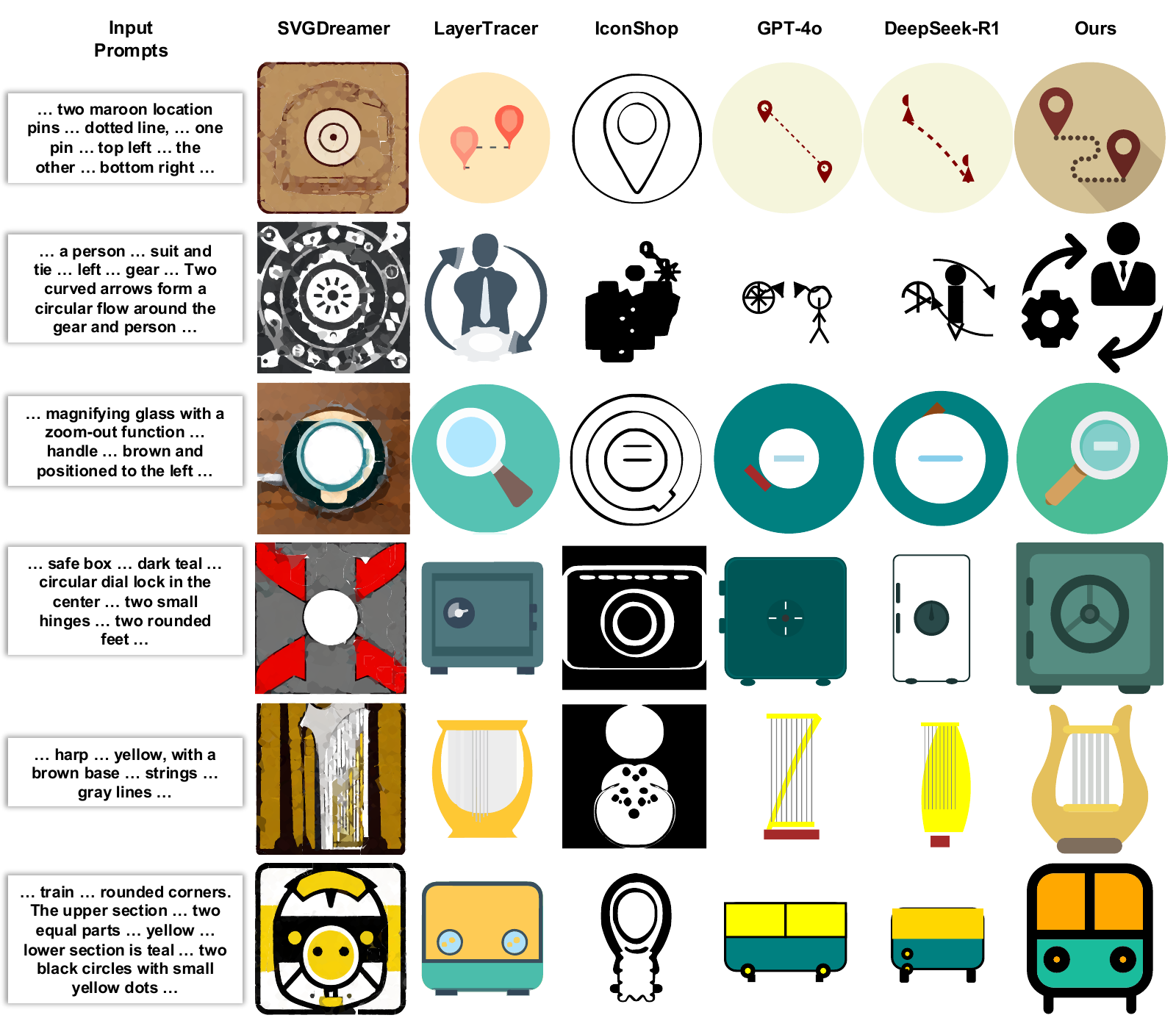}
    \caption{Comparison of generation results. We used instructions from the test dataset as input for all models. As shown, the SVGs generated by our method are better aligned with the instructions and demonstrate superior visual quality.}
    \label{fig:big_compare}
\end{figure*}

The comparison results shown in the Figure \ref{fig:big_compare} indicate that, for various types of generated SVG instructions, our method consistently generates stable and highly usable SVG code. In contrast, image-based methods tend to generate structures that are not well-suited for SVG, with paths that are complex and cumbersome. Furthermore, the resulting SVGs are not smooth or aesthetically pleasing, losing some of the advantages of SVG. Commercial closed-source large language models, currently lacking optimization for SVGs, fail to generate usable SVGs and often produce results that are not visually coordinated or aligned with the generated instructions. From the results in Table \ref{tab:metrics_comparison}, it is evident that our method surpasses existing approaches and models in all metrics. Additionally, our method provides more comprehensive support for SVG instructions, while retaining the essential data storage code format of SVGs, using fewer instructions and smaller file sizes, achieving more efficient and higher-quality SVG generation.


\begin{figure*}
    \centering
    \includegraphics[width=0.91\linewidth]{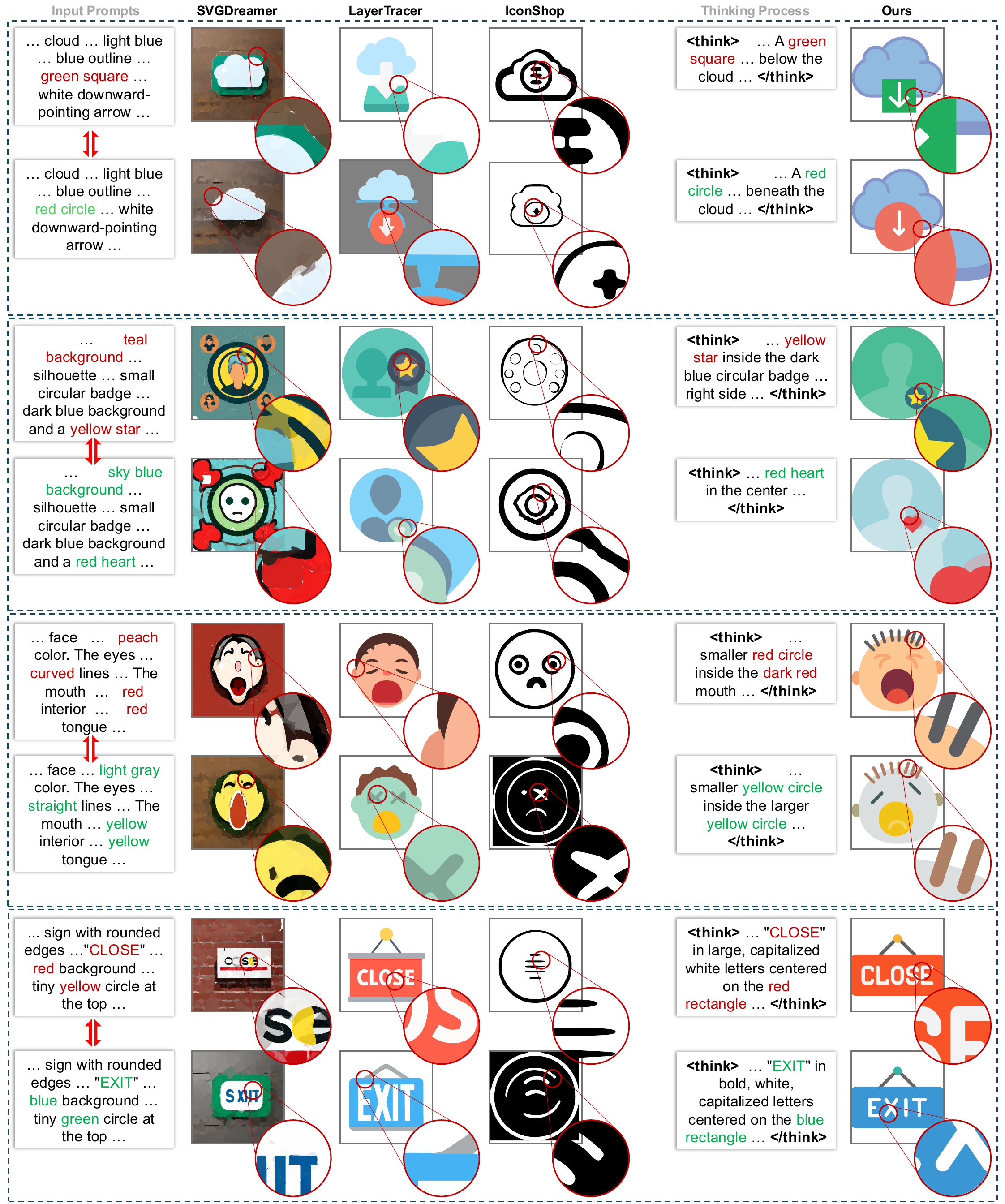}
    \caption{Comparison of test results generated by controllable editing. Each pair forms a controllable editing group, where the input prompt only modifies the description at the corresponding position without altering other sections. Our model can precisely edit the generated SVG results.The thinking process also validates that the model can understand the prompt and grasp the semantic information of SVG primitives. Other models encounter difficulties such as re-generating or failing to understand the prompt. The SVGs produced by our model have sharp, clear edges and are more efficient. In contrast, other SOTA models exhibit flaws and discontinuous mutations to some extent, resulting in a poor visual experience.}
    \label{fig:big_generalize}
\end{figure*}

From the experiments on precise prompt editing in Figure \ref{fig:big_generalize}, it is evident that our model can perform accurate and controllable editing operations. By modifying specific parts of the input instructions, we can make hierarchical, precise adjustments to the relevant areas without affecting the rendering of other parts. In contrast, other baseline models exhibit certain flaws and issues. The details also show that our model generates smoother and more consistent SVGs and better understands the semantic information of SVG instructions. These capabilities are beyond the reach of other models.

\subsection{Ablation Study}

\subsubsection{Ablation Study on Our Method}

First, we conduct an ablation study on our sequential SVG rendering annotation method and the corresponding training approach. The comparison group uses the same model architecture. Since we selected two different weight initialization methods, we also performed experimental tests to examine the impact and effects of these initial weights. Both models were trained using the same hyperparameters based on our training dataset.

The results in Table \ref{tab:ablation_results} show that the performance of the R1 distilled model is worse compared to the Qwen2.5 model. We believe this is mainly because the distilled model primarily learned many specific issues and related thought processes. On one hand, it was not optimized for SVG output; on the other hand, it lost much of the original model’s weight-related SVG knowledge. However, when considering the worse performance of the original weights, the model using our approach performs better. Furthermore, our method significantly improves the model’s ability to generate higher-quality SVGs that align better with textual descriptions. This validates the effectiveness of our approach from an experimental perspective.

\begin{table}
  \caption{Component Analysis. We compared our method with the directly trained method. Our method significantly improved the performance compared to both the original weights and the directly trained method.}
  \label{tab:ablation_results}
  \begin{tabular}{c|c|c|c}
    \toprule
    Methods & FID$\downarrow$ & CLIP$\uparrow$ & FID-CLIP$\downarrow$ \\
    \midrule
    Qwen2.5-7B (zeroshot) & 81.62 & 0.2305 & 65.42 \\
    Qwen2.5-7B (trained) & 41.57 & 0.2345 & 25.92\\
    R1-7B (zeroshot) & 124.87 & 0.2027 & 99.26 \\
    SVGThinker & 34.06 & 0.2765 & 21.08 \\
    \bottomrule
  \end{tabular}
\end{table}

\subsubsection{Ablation Study on Sampling Strategies}

We also conducted an ablation experiment on the hyperparameters of different sampling methods. In the previous experiments, we used greedy search for all models, meaning that the token with the highest probability was selected and output at each step. This method ensures that the output for each specific input is consistent, facilitating comparison and analysis. Given that different sampling methods may still exhibit certain differences, we performed an experimental comparison. We primarily focused on nucleus sampling~\cite{holtzman2019curious} with different hyperparameter selections, as it offers greater novelty and higher-quality generation. In this method, when generating a new token, a set of top-probability candidates is first identified. The range is determined by a pre-set top-p value, and sampling is then performed within the selected range.

\begin{table}
  \caption{Parameter analysis on sampling strategies.}
  \label{tab:sample}
  \begin{tabular}{c|c|c|c}
    \toprule
    Methods & FID$\downarrow$ & CLIP$\uparrow$ & FID-CLIP$\downarrow$ \\
    \midrule
    Greedy Search & 34.06 & 0.2765 & 21.08 \\
    top\_p=0.4, t=0.6 & 33.99 & 0.2766 & 21.43 \\
    top\_p=0.6, t=0.6 & 32.86 & 0.2867 & 20.99 \\
    top\_p=0.8, t=0.6 & 33.87 & 0.2976 & 21.80 \\
    top\_p=1.0, t=0.6 & 33.44 & 0.2854 & 21.08 \\
    top\_p=0.8, t=0.6 & 33.73 & 0.2672 & 21.75 \\
    top\_p=0.8, t=0.8 & 32.79 & 0.2680 & 21.28 \\
    top\_p=0.8, t=1.0 & 33.56 & 0.3182 & 21.50 \\
    \bottomrule
  \end{tabular}
\end{table}

From the results in Table \ref{tab:sample}, we observe that, overall, there is no significant difference in the evaluation data across various sampling hyperparameter settings. Based on the two FID calculation results, the model-generated outputs are relatively robust, consistently aligning with the data distribution of the test set. In terms of CLIP scores, the model demonstrates better alignment with the generation instructions when using different sampling methods, compared to greedy search. This suggests that, in practical applications, we can enhance the generation ability of our model by performing multiple rounds of nucleus sampling and selecting the best results.

\subsection{User Study}

To meet the subjective experience requirements of humans, we conducted a survey with 67 volunteers to collect their opinions and preferences. The volunteers were asked to rank the tools they used from different perspectives, comparing our model with three current state-of-the-art models, as well as including a comparison with commercial models. Considering different evaluation criteria, we asked the volunteers to assess the usability of the methods, the aesthetic quality of the generated content, and the alignment of the generated content with the instructions. Each volunteer used 10 prompts to input into the model and obtained results, then ranked the results. For each volunteer, based on their generated outputs, they were asked to rank all the methods. Finally, we obtained an integrated scoring table and directly performed a weighted average of the scores to obtain the corresponding scores.

\begin{table}
\caption{User study on models comparison.}
\label{tab:user_study}
\begin{tabular}{c|c|c|c}
\toprule
Methods     & Usability$\uparrow$ & Appearance$\uparrow$ & Alignment$\uparrow$ \\ 
\midrule
SVGThinker   & 3.58               & 3.33                & 3.78                     \\ 
SVGDreamer   & 1.25               & 2.69                & 2.58                     \\ 
IconShop    & 3.54               & 3.06                & 3.18                     \\ 
LayerTracer & 2.91               & 3.85                & 3.55                     \\ 
GPT-4o      & 3.76               & 2.16                & 1.91                     \\ 
\bottomrule
\end{tabular}
\end{table}

Results in Table \ref{tab:user_study} indicate that autoregressive models generally provided better user experiences due to inherent interactivity from large language models. Conversely, methods employing longer chains and complex architectures suffered from slower generation speeds and weaker interactivity. Advanced image-generation models achieved higher aesthetic scores, whereas methods without strong image priors were limited by training data, producing less visually appealing SVGs. Notably, our method significantly outperformed others in aligning generated SVGs with user instructions, achieving superior overall results in human preference metrics.

%% file: main.bbl

\begin{thebibliography}{37}


\ifx \showCODEN    \undefined \def \showCODEN     #1{\unskip}     \fi
\ifx \showISBNx    \undefined \def \showISBNx     #1{\unskip}     \fi
\ifx \showISBNxiii \undefined \def \showISBNxiii  #1{\unskip}     \fi
\ifx \showISSN     \undefined \def \showISSN      #1{\unskip}     \fi
\ifx \showLCCN     \undefined \def \showLCCN      #1{\unskip}     \fi
\ifx \shownote     \undefined \def \shownote      #1{#1}          \fi
\ifx \showarticletitle \undefined \def \showarticletitle #1{#1}   \fi
\ifx \showURL      \undefined \def \showURL       {\relax}        \fi
\providecommand\bibfield[2]{#2}
\providecommand\bibinfo[2]{#2}
\providecommand\natexlab[1]{#1}
\providecommand\showeprint[2][]{arXiv:#2}

\bibitem[Achiam et~al\mbox{.}(2023)]%
        {achiam2023gpt}
\bibfield{author}{\bibinfo{person}{Josh Achiam}, \bibinfo{person}{Steven Adler}, \bibinfo{person}{Sandhini Agarwal}, \bibinfo{person}{Lama Ahmad}, \bibinfo{person}{Ilge Akkaya}, \bibinfo{person}{Florencia~Leoni Aleman}, \bibinfo{person}{Diogo Almeida}, \bibinfo{person}{Janko Altenschmidt}, \bibinfo{person}{Sam Altman}, \bibinfo{person}{Shyamal Anadkat}, {et~al\mbox{.}}} \bibinfo{year}{2023}\natexlab{}.
\newblock \showarticletitle{Gpt-4 technical report}.
\newblock \bibinfo{journal}{\emph{arXiv preprint arXiv:2303.08774}} (\bibinfo{year}{2023}).
\newblock


\bibitem[Brown et~al\mbox{.}(2020)]%
        {brown2020language}
\bibfield{author}{\bibinfo{person}{Tom Brown}, \bibinfo{person}{Benjamin Mann}, \bibinfo{person}{Nick Ryder}, \bibinfo{person}{Melanie Subbiah}, \bibinfo{person}{Jared~D Kaplan}, \bibinfo{person}{Prafulla Dhariwal}, \bibinfo{person}{Arvind Neelakantan}, \bibinfo{person}{Pranav Shyam}, \bibinfo{person}{Girish Sastry}, \bibinfo{person}{Amanda Askell}, {et~al\mbox{.}}} \bibinfo{year}{2020}\natexlab{}.
\newblock \showarticletitle{Language models are few-shot learners}.
\newblock \bibinfo{journal}{\emph{Advances in neural information processing systems}}  \bibinfo{volume}{33} (\bibinfo{year}{2020}), \bibinfo{pages}{1877--1901}.
\newblock


\bibitem[Carlier et~al\mbox{.}(2020)]%
        {carlier2020deepsvg}
\bibfield{author}{\bibinfo{person}{Alexandre Carlier}, \bibinfo{person}{Martin Danelljan}, \bibinfo{person}{Alexandre Alahi}, {and} \bibinfo{person}{Radu Timofte}.} \bibinfo{year}{2020}\natexlab{}.
\newblock \showarticletitle{Deepsvg: A hierarchical generative network for vector graphics animation}.
\newblock \bibinfo{journal}{\emph{Advances in Neural Information Processing Systems}}  \bibinfo{volume}{33} (\bibinfo{year}{2020}), \bibinfo{pages}{16351--16361}.
\newblock


\bibitem[Chen et~al\mbox{.}(2024)]%
        {chen2024expanding}
\bibfield{author}{\bibinfo{person}{Zhe Chen}, \bibinfo{person}{Weiyun Wang}, \bibinfo{person}{Yue Cao}, \bibinfo{person}{Yangzhou Liu}, \bibinfo{person}{Zhangwei Gao}, \bibinfo{person}{Erfei Cui}, \bibinfo{person}{Jinguo Zhu}, \bibinfo{person}{Shenglong Ye}, \bibinfo{person}{Hao Tian}, \bibinfo{person}{Zhaoyang Liu}, {et~al\mbox{.}}} \bibinfo{year}{2024}\natexlab{}.
\newblock \showarticletitle{Expanding Performance Boundaries of Open-Source Multimodal Models with Model, Data, and Test-Time Scaling}.
\newblock \bibinfo{journal}{\emph{arXiv preprint arXiv:2412.05271}} (\bibinfo{year}{2024}).
\newblock


\bibitem[Condino(2022)]%
        {svgkaggle}
\bibfield{author}{\bibinfo{person}{Victor Condino}.} \bibinfo{year}{2022}\natexlab{}.
\newblock \bibinfo{title}{Large collection of labelled SVG icon graphics in a variety of styles}.
\newblock \bibinfo{howpublished}{\url{https://www.kaggle.com/datasets/victorcondino/svgicons/}}.
\newblock


\bibitem[Contributors(2023)]%
        {2023lmdeploy}
\bibfield{author}{\bibinfo{person}{LMDeploy Contributors}.} \bibinfo{year}{2023}\natexlab{}.
\newblock \bibinfo{title}{LMDeploy: A Toolkit for Compressing, Deploying, and Serving LLM}.
\newblock \bibinfo{howpublished}{\url{https://github.com/InternLM/lmdeploy}}.
\newblock


\bibitem[Cortex(2023)]%
        {vtracer}
\bibfield{author}{\bibinfo{person}{Vision Cortex}.} \bibinfo{year}{2023}\natexlab{}.
\newblock \bibinfo{title}{Raster to Vector Graphics Converter built on top of visioncortex}.
\newblock \bibinfo{howpublished}{\url{https://github.com/visioncortex/vtracer}}.
\newblock


\bibitem[Guo et~al\mbox{.}(2025)]%
        {guo2025deepseek}
\bibfield{author}{\bibinfo{person}{Daya Guo}, \bibinfo{person}{Dejian Yang}, \bibinfo{person}{Haowei Zhang}, \bibinfo{person}{Junxiao Song}, \bibinfo{person}{Ruoyu Zhang}, \bibinfo{person}{Runxin Xu}, \bibinfo{person}{Qihao Zhu}, \bibinfo{person}{Shirong Ma}, \bibinfo{person}{Peiyi Wang}, \bibinfo{person}{Xiao Bi}, {et~al\mbox{.}}} \bibinfo{year}{2025}\natexlab{}.
\newblock \showarticletitle{Deepseek-r1: Incentivizing reasoning capability in llms via reinforcement learning}.
\newblock \bibinfo{journal}{\emph{arXiv preprint arXiv:2501.12948}} (\bibinfo{year}{2025}).
\newblock


\bibitem[Heusel et~al\mbox{.}(2017)]%
        {heusel2017gans}
\bibfield{author}{\bibinfo{person}{Martin Heusel}, \bibinfo{person}{Hubert Ramsauer}, \bibinfo{person}{Thomas Unterthiner}, \bibinfo{person}{Bernhard Nessler}, {and} \bibinfo{person}{Sepp Hochreiter}.} \bibinfo{year}{2017}\natexlab{}.
\newblock \showarticletitle{Gans trained by a two time-scale update rule converge to a local nash equilibrium}.
\newblock \bibinfo{journal}{\emph{Advances in neural information processing systems}}  \bibinfo{volume}{30} (\bibinfo{year}{2017}).
\newblock


\bibitem[Ho et~al\mbox{.}(2020)]%
        {ho2020denoising}
\bibfield{author}{\bibinfo{person}{Jonathan Ho}, \bibinfo{person}{Ajay Jain}, {and} \bibinfo{person}{Pieter Abbeel}.} \bibinfo{year}{2020}\natexlab{}.
\newblock \showarticletitle{Denoising diffusion probabilistic models}.
\newblock \bibinfo{journal}{\emph{Advances in neural information processing systems}}  \bibinfo{volume}{33} (\bibinfo{year}{2020}), \bibinfo{pages}{6840--6851}.
\newblock


\bibitem[Holtzman et~al\mbox{.}(2019)]%
        {holtzman2019curious}
\bibfield{author}{\bibinfo{person}{Ari Holtzman}, \bibinfo{person}{Jan Buys}, \bibinfo{person}{Li Du}, \bibinfo{person}{Maxwell Forbes}, {and} \bibinfo{person}{Yejin Choi}.} \bibinfo{year}{2019}\natexlab{}.
\newblock \showarticletitle{The curious case of neural text degeneration}.
\newblock \bibinfo{journal}{\emph{arXiv preprint arXiv:1904.09751}} (\bibinfo{year}{2019}).
\newblock


\bibitem[Jain et~al\mbox{.}(2023)]%
        {jain2023vectorfusion}
\bibfield{author}{\bibinfo{person}{Ajay Jain}, \bibinfo{person}{Amber Xie}, {and} \bibinfo{person}{Pieter Abbeel}.} \bibinfo{year}{2023}\natexlab{}.
\newblock \showarticletitle{Vectorfusion: Text-to-svg by abstracting pixel-based diffusion models}. In \bibinfo{booktitle}{\emph{Proceedings of the IEEE/CVF Conference on Computer Vision and Pattern Recognition}}. \bibinfo{pages}{1911--1920}.
\newblock


\bibitem[Kaplan et~al\mbox{.}(2020)]%
        {kaplan2020scaling}
\bibfield{author}{\bibinfo{person}{Jared Kaplan}, \bibinfo{person}{Sam McCandlish}, \bibinfo{person}{Tom Henighan}, \bibinfo{person}{Tom~B Brown}, \bibinfo{person}{Benjamin Chess}, \bibinfo{person}{Rewon Child}, \bibinfo{person}{Scott Gray}, \bibinfo{person}{Alec Radford}, \bibinfo{person}{Jeffrey Wu}, {and} \bibinfo{person}{Dario Amodei}.} \bibinfo{year}{2020}\natexlab{}.
\newblock \showarticletitle{Scaling laws for neural language models}.
\newblock \bibinfo{journal}{\emph{arXiv preprint arXiv:2001.08361}} (\bibinfo{year}{2020}).
\newblock


\bibitem[Labs(2024)]%
        {flux2024}
\bibfield{author}{\bibinfo{person}{Black~Forest Labs}.} \bibinfo{year}{2024}\natexlab{}.
\newblock \bibinfo{title}{FLUX}.
\newblock \bibinfo{howpublished}{\url{https://github.com/black-forest-labs/flux}}.
\newblock


\bibitem[Li et~al\mbox{.}(2020)]%
        {Li:2020:DVG}
\bibfield{author}{\bibinfo{person}{Tzu-Mao Li}, \bibinfo{person}{Michal Luk\'{a}\v{c}}, \bibinfo{person}{Gharbi Micha\"{e}l}, {and} \bibinfo{person}{Jonathan Ragan-Kelley}.} \bibinfo{year}{2020}\natexlab{}.
\newblock \showarticletitle{Differentiable Vector Graphics Rasterization for Editing and Learning}.
\newblock \bibinfo{journal}{\emph{ACM Trans. Graph. (Proc. SIGGRAPH Asia)}} \bibinfo{volume}{39}, \bibinfo{number}{6} (\bibinfo{year}{2020}), \bibinfo{pages}{193:1--193:15}.
\newblock


\bibitem[Lin et~al\mbox{.}(2024)]%
        {lin2024awq}
\bibfield{author}{\bibinfo{person}{Ji Lin}, \bibinfo{person}{Jiaming Tang}, \bibinfo{person}{Haotian Tang}, \bibinfo{person}{Shang Yang}, \bibinfo{person}{Wei-Ming Chen}, \bibinfo{person}{Wei-Chen Wang}, \bibinfo{person}{Guangxuan Xiao}, \bibinfo{person}{Xingyu Dang}, \bibinfo{person}{Chuang Gan}, {and} \bibinfo{person}{Song Han}.} \bibinfo{year}{2024}\natexlab{}.
\newblock \showarticletitle{Awq: Activation-aware weight quantization for on-device llm compression and acceleration}.
\newblock \bibinfo{journal}{\emph{Proceedings of Machine Learning and Systems}}  \bibinfo{volume}{6} (\bibinfo{year}{2024}), \bibinfo{pages}{87--100}.
\newblock


\bibitem[Liu et~al\mbox{.}(2023)]%
        {liu2023visual}
\bibfield{author}{\bibinfo{person}{Haotian Liu}, \bibinfo{person}{Chunyuan Li}, \bibinfo{person}{Qingyang Wu}, {and} \bibinfo{person}{Yong~Jae Lee}.} \bibinfo{year}{2023}\natexlab{}.
\newblock \showarticletitle{Visual instruction tuning}.
\newblock \bibinfo{journal}{\emph{Advances in neural information processing systems}}  \bibinfo{volume}{36} (\bibinfo{year}{2023}), \bibinfo{pages}{34892--34916}.
\newblock


\bibitem[Lopes et~al\mbox{.}(2019)]%
        {lopes2019learned}
\bibfield{author}{\bibinfo{person}{Raphael~Gontijo Lopes}, \bibinfo{person}{David Ha}, \bibinfo{person}{Douglas Eck}, {and} \bibinfo{person}{Jonathon Shlens}.} \bibinfo{year}{2019}\natexlab{}.
\newblock \showarticletitle{A learned representation for scalable vector graphics}. In \bibinfo{booktitle}{\emph{Proceedings of the IEEE/CVF International Conference on Computer Vision}}. \bibinfo{pages}{7930--7939}.
\newblock


\bibitem[Loshchilov and Hutter(2016)]%
        {loshchilov2016sgdr}
\bibfield{author}{\bibinfo{person}{Ilya Loshchilov} {and} \bibinfo{person}{Frank Hutter}.} \bibinfo{year}{2016}\natexlab{}.
\newblock \showarticletitle{Sgdr: Stochastic gradient descent with warm restarts}.
\newblock \bibinfo{journal}{\emph{arXiv preprint arXiv:1608.03983}} (\bibinfo{year}{2016}).
\newblock


\bibitem[Loshchilov and Hutter(2017)]%
        {loshchilov2017decoupled}
\bibfield{author}{\bibinfo{person}{Ilya Loshchilov} {and} \bibinfo{person}{Frank Hutter}.} \bibinfo{year}{2017}\natexlab{}.
\newblock \showarticletitle{Decoupled weight decay regularization}.
\newblock \bibinfo{journal}{\emph{arXiv preprint arXiv:1711.05101}} (\bibinfo{year}{2017}).
\newblock


\bibitem[Radford et~al\mbox{.}(2021)]%
        {radford2021learning}
\bibfield{author}{\bibinfo{person}{Alec Radford}, \bibinfo{person}{Jong~Wook Kim}, \bibinfo{person}{Chris Hallacy}, \bibinfo{person}{Aditya Ramesh}, \bibinfo{person}{Gabriel Goh}, \bibinfo{person}{Sandhini Agarwal}, \bibinfo{person}{Girish Sastry}, \bibinfo{person}{Amanda Askell}, \bibinfo{person}{Pamela Mishkin}, \bibinfo{person}{Jack Clark}, {et~al\mbox{.}}} \bibinfo{year}{2021}\natexlab{}.
\newblock \showarticletitle{Learning transferable visual models from natural language supervision}. In \bibinfo{booktitle}{\emph{International conference on machine learning}}. PmLR, \bibinfo{pages}{8748--8763}.
\newblock


\bibitem[Radford et~al\mbox{.}(2019)]%
        {radford2019language}
\bibfield{author}{\bibinfo{person}{Alec Radford}, \bibinfo{person}{Jeffrey Wu}, \bibinfo{person}{Rewon Child}, \bibinfo{person}{David Luan}, \bibinfo{person}{Dario Amodei}, \bibinfo{person}{Ilya Sutskever}, {et~al\mbox{.}}} \bibinfo{year}{2019}\natexlab{}.
\newblock \showarticletitle{Language models are unsupervised multitask learners}.
\newblock \bibinfo{journal}{\emph{OpenAI blog}} \bibinfo{volume}{1}, \bibinfo{number}{8} (\bibinfo{year}{2019}), \bibinfo{pages}{9}.
\newblock


\bibitem[Rodriguez et~al\mbox{.}(2023)]%
        {rodriguez2023starvector}
\bibfield{author}{\bibinfo{person}{Juan~A Rodriguez}, \bibinfo{person}{Shubham Agarwal}, \bibinfo{person}{Issam~H Laradji}, \bibinfo{person}{Pau Rodriguez}, \bibinfo{person}{David Vazquez}, \bibinfo{person}{Christopher Pal}, {and} \bibinfo{person}{Marco Pedersoli}.} \bibinfo{year}{2023}\natexlab{}.
\newblock \showarticletitle{Starvector: Generating scalable vector graphics code from images}.
\newblock \bibinfo{journal}{\emph{arXiv preprint arXiv:2312.11556}} (\bibinfo{year}{2023}).
\newblock


\bibitem[Rombach et~al\mbox{.}(2022)]%
        {rombach2022high}
\bibfield{author}{\bibinfo{person}{Robin Rombach}, \bibinfo{person}{Andreas Blattmann}, \bibinfo{person}{Dominik Lorenz}, \bibinfo{person}{Patrick Esser}, {and} \bibinfo{person}{Bj{\"o}rn Ommer}.} \bibinfo{year}{2022}\natexlab{}.
\newblock \showarticletitle{High-resolution image synthesis with latent diffusion models}. In \bibinfo{booktitle}{\emph{Proceedings of the IEEE/CVF conference on computer vision and pattern recognition}}. \bibinfo{pages}{10684--10695}.
\newblock


\bibitem[Song et~al\mbox{.}(2025)]%
        {song2025layertracer}
\bibfield{author}{\bibinfo{person}{Yiren Song}, \bibinfo{person}{Danze Chen}, {and} \bibinfo{person}{Mike~Zheng Shou}.} \bibinfo{year}{2025}\natexlab{}.
\newblock \showarticletitle{LayerTracer: Cognitive-Aligned Layered SVG Synthesis via Diffusion Transformer}.
\newblock \bibinfo{journal}{\emph{arXiv preprint arXiv:2502.01105}} (\bibinfo{year}{2025}).
\newblock


\bibitem[SVGRepo(2016)]%
        {svgrepo}
\bibfield{author}{\bibinfo{person}{SVGRepo}.} \bibinfo{year}{2016}\natexlab{}.
\newblock \bibinfo{title}{Open-licensed svg vector and icons}.
\newblock \bibinfo{howpublished}{\url{https://www.svgrepo.com/}}.
\newblock


\bibitem[Szegedy et~al\mbox{.}(2016)]%
        {szegedy2016rethinking}
\bibfield{author}{\bibinfo{person}{Christian Szegedy}, \bibinfo{person}{Vincent Vanhoucke}, \bibinfo{person}{Sergey Ioffe}, \bibinfo{person}{Jon Shlens}, {and} \bibinfo{person}{Zbigniew Wojna}.} \bibinfo{year}{2016}\natexlab{}.
\newblock \showarticletitle{Rethinking the inception architecture for computer vision}. In \bibinfo{booktitle}{\emph{Proceedings of the IEEE conference on computer vision and pattern recognition}}. \bibinfo{pages}{2818--2826}.
\newblock


\bibitem[Team(2024)]%
        {qwen2.5}
\bibfield{author}{\bibinfo{person}{Qwen Team}.} \bibinfo{year}{2024}\natexlab{}.
\newblock \bibinfo{title}{Qwen2.5: A Party of Foundation Models}.
\newblock
\urldef\tempurl%
\url{https://qwenlm.github.io/blog/qwen2.5/}
\showURL{%
\tempurl}


\bibitem[Vaswani et~al\mbox{.}(2017)]%
        {vaswani2017attention}
\bibfield{author}{\bibinfo{person}{Ashish Vaswani}, \bibinfo{person}{Noam Shazeer}, \bibinfo{person}{Niki Parmar}, \bibinfo{person}{Jakob Uszkoreit}, \bibinfo{person}{Llion Jones}, \bibinfo{person}{Aidan~N Gomez}, \bibinfo{person}{{\L}ukasz Kaiser}, {and} \bibinfo{person}{Illia Polosukhin}.} \bibinfo{year}{2017}\natexlab{}.
\newblock \showarticletitle{Attention is all you need}.
\newblock \bibinfo{journal}{\emph{Advances in neural information processing systems}}  \bibinfo{volume}{30} (\bibinfo{year}{2017}).
\newblock


\bibitem[Wei et~al\mbox{.}(2022)]%
        {wei2022chain}
\bibfield{author}{\bibinfo{person}{Jason Wei}, \bibinfo{person}{Xuezhi Wang}, \bibinfo{person}{Dale Schuurmans}, \bibinfo{person}{Maarten Bosma}, \bibinfo{person}{Fei Xia}, \bibinfo{person}{Ed Chi}, \bibinfo{person}{Quoc~V Le}, \bibinfo{person}{Denny Zhou}, {et~al\mbox{.}}} \bibinfo{year}{2022}\natexlab{}.
\newblock \showarticletitle{Chain-of-thought prompting elicits reasoning in large language models}.
\newblock \bibinfo{journal}{\emph{Advances in neural information processing systems}}  \bibinfo{volume}{35} (\bibinfo{year}{2022}), \bibinfo{pages}{24824--24837}.
\newblock


\bibitem[Wu et~al\mbox{.}(2024)]%
        {wu2024chat2svg}
\bibfield{author}{\bibinfo{person}{Ronghuan Wu}, \bibinfo{person}{Wanchao Su}, {and} \bibinfo{person}{Jing Liao}.} \bibinfo{year}{2024}\natexlab{}.
\newblock \showarticletitle{Chat2SVG: Vector Graphics Generation with Large Language Models and Image Diffusion Models}.
\newblock \bibinfo{journal}{\emph{arXiv preprint arXiv:2411.16602}} (\bibinfo{year}{2024}).
\newblock


\bibitem[Wu et~al\mbox{.}(2023)]%
        {wu2023iconshop}
\bibfield{author}{\bibinfo{person}{Ronghuan Wu}, \bibinfo{person}{Wanchao Su}, \bibinfo{person}{Kede Ma}, {and} \bibinfo{person}{Jing Liao}.} \bibinfo{year}{2023}\natexlab{}.
\newblock \showarticletitle{Iconshop: Text-guided vector icon synthesis with autoregressive transformers}.
\newblock \bibinfo{journal}{\emph{ACM Transactions on Graphics (TOG)}} \bibinfo{volume}{42}, \bibinfo{number}{6} (\bibinfo{year}{2023}), \bibinfo{pages}{1--14}.
\newblock


\bibitem[Xing et~al\mbox{.}(2024a)]%
        {xing2024empowering}
\bibfield{author}{\bibinfo{person}{Ximing Xing}, \bibinfo{person}{Juncheng Hu}, \bibinfo{person}{Guotao Liang}, \bibinfo{person}{Jing Zhang}, \bibinfo{person}{Dong Xu}, {and} \bibinfo{person}{Qian Yu}.} \bibinfo{year}{2024}\natexlab{a}.
\newblock \showarticletitle{Empowering LLMs to Understand and Generate Complex Vector Graphics}.
\newblock \bibinfo{journal}{\emph{arXiv preprint arXiv:2412.11102}} (\bibinfo{year}{2024}).
\newblock


\bibitem[Xing et~al\mbox{.}(2024b)]%
        {xing2024svgdreamer}
\bibfield{author}{\bibinfo{person}{Ximing Xing}, \bibinfo{person}{Haitao Zhou}, \bibinfo{person}{Chuang Wang}, \bibinfo{person}{Jing Zhang}, \bibinfo{person}{Dong Xu}, {and} \bibinfo{person}{Qian Yu}.} \bibinfo{year}{2024}\natexlab{b}.
\newblock \showarticletitle{Svgdreamer: Text guided svg generation with diffusion model}. In \bibinfo{booktitle}{\emph{Proceedings of the IEEE/CVF Conference on Computer Vision and Pattern Recognition}}. \bibinfo{pages}{4546--4555}.
\newblock


\bibitem[Zhang et~al\mbox{.}(2023)]%
        {zhang2023multimodal}
\bibfield{author}{\bibinfo{person}{Zhuosheng Zhang}, \bibinfo{person}{Aston Zhang}, \bibinfo{person}{Mu Li}, \bibinfo{person}{Hai Zhao}, \bibinfo{person}{George Karypis}, {and} \bibinfo{person}{Alex Smola}.} \bibinfo{year}{2023}\natexlab{}.
\newblock \showarticletitle{Multimodal chain-of-thought reasoning in language models}.
\newblock \bibinfo{journal}{\emph{arXiv preprint arXiv:2302.00923}} (\bibinfo{year}{2023}).
\newblock


\bibitem[Zhao et~al\mbox{.}(2024)]%
        {zhao2024vector}
\bibfield{author}{\bibinfo{person}{Zhongyin Zhao}, \bibinfo{person}{Ye Chen}, \bibinfo{person}{Zhangli Hu}, \bibinfo{person}{Xuanhong Chen}, {and} \bibinfo{person}{Bingbing Ni}.} \bibinfo{year}{2024}\natexlab{}.
\newblock \showarticletitle{Vector Graphics Generation via Mutually Impulsed Dual-domain Diffusion}. In \bibinfo{booktitle}{\emph{Proceedings of the IEEE/CVF Conference on Computer Vision and Pattern Recognition}}. \bibinfo{pages}{4420--4428}.
\newblock


\bibitem[Zheng et~al\mbox{.}(2024)]%
        {zheng2024llamafactory}
\bibfield{author}{\bibinfo{person}{Yaowei Zheng}, \bibinfo{person}{Richong Zhang}, \bibinfo{person}{Junhao Zhang}, \bibinfo{person}{Yanhan Ye}, \bibinfo{person}{Zheyan Luo}, \bibinfo{person}{Zhangchi Feng}, {and} \bibinfo{person}{Yongqiang Ma}.} \bibinfo{year}{2024}\natexlab{}.
\newblock \showarticletitle{Llamafactory: Unified efficient fine-tuning of 100+ language models}.
\newblock \bibinfo{journal}{\emph{arXiv preprint arXiv:2403.13372}} (\bibinfo{year}{2024}).
\newblock


\end{thebibliography}
